\documentclass{article} 
\usepackage[final]{colm2026_conference}

\usepackage{microtype}
\usepackage{graphicx}
\usepackage{subcaption}
\usepackage{booktabs}

\usepackage{hyperref}
\usepackage{url}

\usepackage{amsmath}
\usepackage{amssymb}
\usepackage{amsfonts}

\usepackage[utf8]{inputenc}
\usepackage[T1]{fontenc}
\usepackage{nicefrac}
\usepackage{xcolor}
\usepackage{xspace}
\usepackage{enumitem}
\usepackage{multirow}
\usepackage{pifont}
\usepackage{float}
\usepackage{rotating}
\usepackage{wrapfig}

\usepackage[capitalize,noabbrev]{cleveref}

\usepackage[most]{tcolorbox}
\tcbuselibrary{breakable}

\usepackage{algorithm}
\usepackage{algorithmic}

\usepackage{lineno}

\definecolor{darkblue}{rgb}{0, 0, 0.5}
\hypersetup{colorlinks=true, citecolor=darkblue, linkcolor=darkblue, urlcolor=darkblue}

\newcommand{\modelname}{\textbf{v1}\xspace}

\newcommand{\dataname}{\textbf{v1g}\xspace}

\definecolor{customgreen}{HTML}{04BC7D}
\definecolor{customred}{HTML}{EF3333}
\newcommand{\cmark}{\textcolor{customgreen}{\ding{51}}}
\newcommand{\xmark}{\textcolor{customred}{\ding{55}}}
\newcommand{\versus}{\textit{vs.}\xspace}
\newcommand{\eg}{\emph{e.g.}\xspace}

\title{v1: Learning to Point Visual Tokens\\for Multimodal Grounded Reasoning}

\author{Jiwan Chung$^{1}$\thanks{Equal contribution.} \qquad Junhyeok Kim$^{1}$\footnotemark[1] \qquad Siyeol Kim$^{1}$\\
\textbf{Jaeyoung Lee$^{1}$ \qquad Min Soo Kim$^{1}$ \qquad Youngjae Yu$^{2}$}
\\\\$^{1}$Yonsei University \qquad $^{2}$Seoul National University \\
\texttt{jiwan.chung.research@gmail.com} \qquad \texttt{junhyeok@yonsei.ac.kr}
}

\ifcolmfinal\fi

\begin{document}

\ifcolmsubmission
\linenumbers
\fi

\maketitle


\begin{abstract}
When thinking with images, humans rarely rely on a single glance: they revisit visual evidence while reasoning. In contrast, most Multimodal Language Models encode an image once to key-value cache and then reason purely in text, making it hard to re-ground intermediate steps. We empirically confirm this: as reasoning chains lengthen, models progressively lose focus on relevant regions.
We introduce \modelname, a lightweight extension for active visual referencing via point-and-copy: the model selects relevant image patches and copies their embeddings back into the reasoning stream. Crucially, our point-and-copy mechanism retrieves patches using their semantic representations as keys, ensuring perceptual evidence remains aligned with the reasoning space.
To train this behavior, we build \dataname, a dataset of 300K multimodal reasoning traces with interleaved grounding annotations. Across multimodal mathematical reasoning benchmarks, \modelname consistently outperforms comparable baselines. We release our code, model, and data.
\end{abstract}

\section{Introduction}
\label{sec:intro}

Multimodal mathematical problem solving often requires repeated access to diagrams: intermediate steps depend on localized cues (e.g., angles, tangency points, symmetries) that are revisited as reasoning unfolds. Cognitive studies report similar visual revisitation~\citep{Cox1999RepresentationCE, Brun2016DesigningWS, Chu2017DiagramsBS, Kozhevnikov01032002}.

Recent advances in Multimodal Large Language Models (MLLMs) \citep{liu2023llava, qwen25vl, internvl25} have shifted focus toward complex multi-step reasoning \citep{xu2025llavacotletvisionlanguage, yao2024mulberry, sun2025tvc, huang2025v-r1} rather than direct recognition. Mathematical reasoning \citep{lu2024mathvista, zhang2024mathverse, wang2024measuring} is a standard benchmark for this capability; it requires long-horizon grounding and unambiguous solutions, making it a natural testbed for grounded multimodal inference.

However, current MLLMs process images only once at the start and, due to causal masking, thereafter reason mainly over the frozen key--value cache of visual embeddings. This limits their ability to actively revisit visual context as inference unfolds.
In practice, this constraint manifests as two forms of \textit{visual grounding decay}. First, attention to all image tokens steadily weakens as reasoning chains extend. Second, even the relative weight on relevant tokens declines, reducing the model's ability to focus on the most informative regions (\cref{sec:prelim_exp}).
These effects highlight the need for mechanisms that let models explicitly re-access visual information to keep reasoning grounded in the input.

\begin{figure}[t]
\begin{center}
\includegraphics[width=1\textwidth]{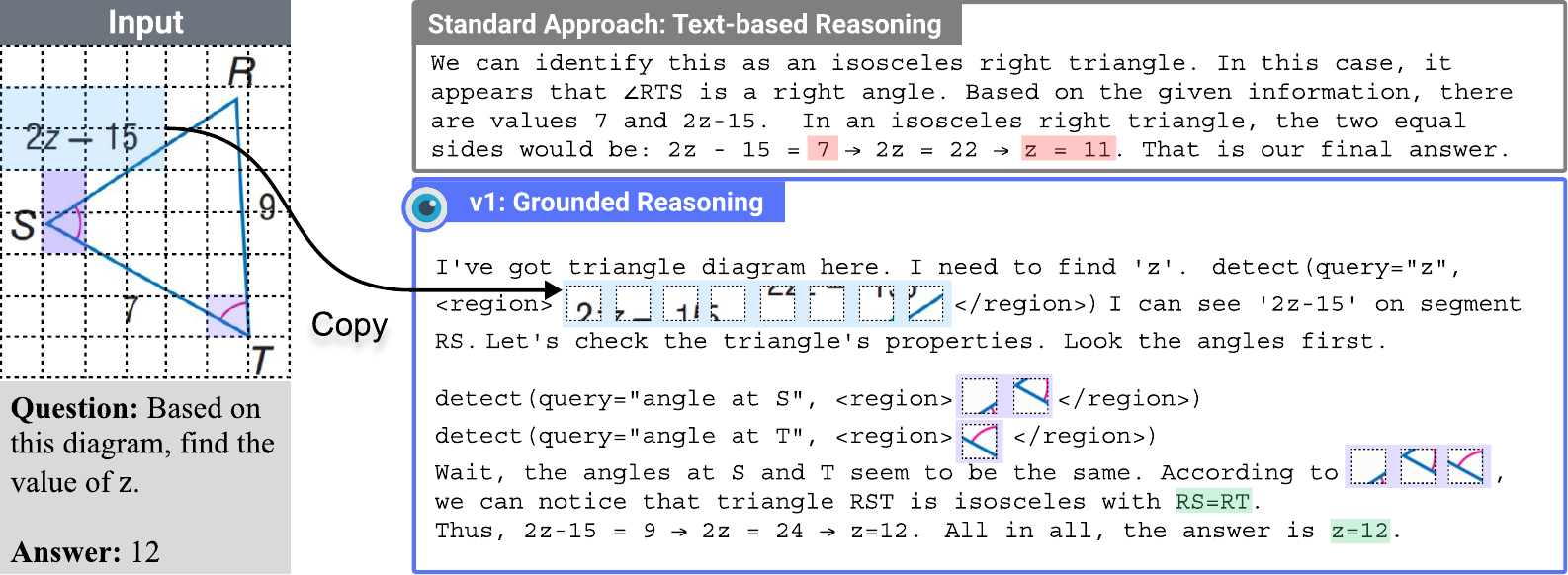}
\end{center}
\caption{\textbf{Pure text-based reasoning \versus \modelname during inference.}
\modelname can actively re-access visual context by point-and-copying relevant image regions throughout reasoning process.}
  \label{fig:method}
\end{figure}


We propose \modelname\footnote{Code: \url{https://github.com/jun297/v1}. Model: \url{https://huggingface.co/kjunh/v1-7B}. Data (\dataname): \url{https://huggingface.co/datasets/kjunh/v1g}. Project page: \url{https://jun297.github.io/v1/}.}, a simple yet effective extension that equips MLLMs with a \textit{point-and-copy} mechanism for dynamically referencing input visual tokens during multimodal reasoning (\cref{fig:method}). Specifically, we augment the model with an additional pointing head that outputs a probability distribution over the input image token positions, alongside the standard vocabulary logits. When an image token is selected, its embedding is copied and injected as the next-step input, allowing the model to dynamically retrieve and reuse visual information during generation.

Our approach is readily compatible with popular MLLM architectures~\citep{liu2023llava,qwen2vl,internvl25} that operate on continuous image embeddings. Unlike methods that attempt to generate new image tokens \citep{team2024chameleon} which require full-scale pretraining for image generation capacity, our method simply reuses existing input embeddings through pointing and copying. The only additional parameters are lightweight linear heads, incurring minimal computational overhead.

To train \modelname, we construct \dataname, a dataset of 300K multimodal mathematical reasoning paths with interleaved grounding annotations, where each reasoning step is explicitly linked to a corresponding image region. The construction pipeline comprises three stages: (1) oversampling diverse reasoning traces from an MLLM, (2) extracting visual queries and retrieval steps from the traces using an LLM-guided decomposition process, and (3) grounding each visual reference by associating it with a bounding box in the input image. The pipeline is fully automated, leveraging the generative and interpretive capabilities of LLMs to produce high-quality, grounded reasoning trajectories at scale.

We evaluate \modelname on three established multimodal mathematical reasoning benchmarks: MathVista~\citep{lu2024mathvista}, MathVision~\citep{wang2024measuring}, and MathVerse~\citep{zhang2024mathverse}, following prior work~\citep{yao2024mulberry, sun2025tvc, huang2025v-r1}.
\modelname demonstrates strong performance across all benchmarks, outperforming existing models of comparable scale and approaching the capabilities of much larger models, particularly on tasks requiring precise visual grounding and iterative reference to localized regions. These results suggest that dynamic access of visual input at inference time can improve multimodal mathematical reasoning capabilities.

Our contributions are:
\begin{itemize}[leftmargin=*,nosep]
    \item \textbf{\modelname model:} a lightweight MLLM extension that helps mitigate visual grounding decay through a point-and-copy mechanism, enabling dynamic visual reference.
    \item \textbf{\dataname dataset:} a large-scale training set with 300K multimodal mathematical reasoning traces and fine-grained visual grounding.
    \item \textbf{Empirical findings:} extensive experiments and ablations on multimodal mathematical reasoning benchmarks, showing that dynamic visual reference and the point-and-copy design both mitigate visual grounding decay and lead to better multimodal reasoning.  
\end{itemize}

\begin{figure}
  \centering
  \includegraphics[width=1\textwidth]{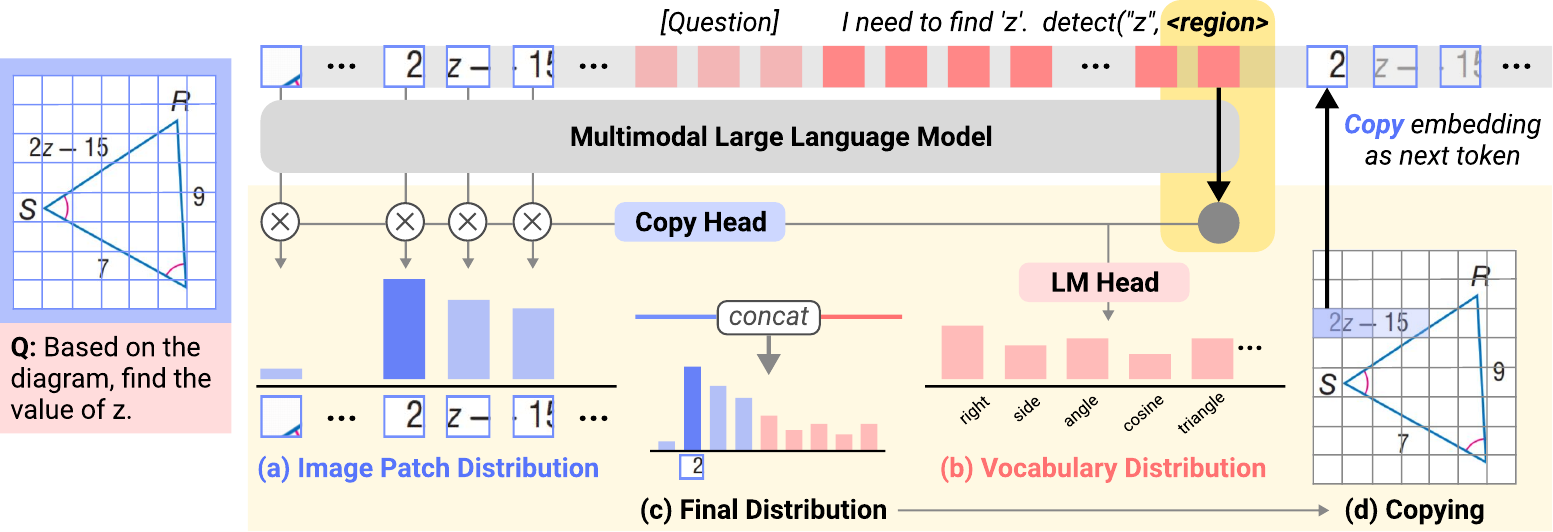}
  \caption{\textbf{Inference process of \modelname.} At each step, the MLLM encodes the multimodal context and generation history into token representations. For the last token (e.g., "<region>"), (a) a copy head projects its representation and computes logits against image patch embeddings, (b) a language head produces logits over the vocabulary, and (c) the two are concatenated to form the final distribution. If a patch is chosen, its embedding is copied as the next token input, enabling \modelname to reference image regions one patch at a time.}
  \label{fig:impl}
\end{figure}

\section{Related Work}
\label{sec:rel}

\subsection{Reasoning in Large Language Models}
\label{subsec:rel_reason}


\paragraph{Reasoning in text-only large language models.} 
Recent advances have significantly improved reasoning in text-only LLMs. OpenAI's o1 model~\citep{openai2024o1card} had achieved state-of-the-art performance on mathematical benchmarks~\citep{lightman2023let, cobbe2021gsm8k}, motivating follow-up work that induces stronger reasoning through reinforcement learning and reflective Chain-of-Thought~\citep{deepseekai2025deepseekr1}, as well as inference-time scaling~\citep{muennighoff2025s1}. While effective in text-only settings, extending these approaches to multimodal reasoning remains challenging.

\paragraph{Reasoning in multimodal large language models.}

Multimodal reasoning poses challenges beyond text-only inference, requiring both visual perception and the integration of visual inputs into multi-step reasoning. Early approaches~\citep{liu2023llava, chen2024spatialvlm, zhang2024mcot, yang2023mmreact} typically convert images into textual descriptions for downstream tasks. More recently, inspired by Chain-of-Thought prompting in LLMs, models such as LLaVA-CoT~\citep{xu2025llavacotletvisionlanguage}, TVC~\citep{sun2025tvc}, and Mulberry~\citep{yao2024mulberry}, among others~\citep{huang2025v-r1, deng2025openvlthinker, meng2025mmeureka, gcot, wang2025sotalessmctsguidedsample}, extend CoT reasoning to multimodal settings and achieve strong results on benchmarks such as MathVista~\citep{lu2024mathvista} and MathVision~\citep{wang2024measuring}. However, these models treat visual inputs as fixed context and perform reasoning entirely in the text space, lacking explicit mechanisms to re-access visual representations.

\subsection{Implementing Visual Reference}
\label{subsec:rel_refer}
A growing body of work equips MLLMs with mechanisms to interact with images more explicitly during reasoning, beyond treating the visual input as a single, static context. Existing approaches span a range of design choices, including predicting structured spatial references (e.g., boxes or points) to retrieve localized evidence~\citep{visualprogramming,wu2023vstar,chung2025teachingmetricdistancediscrete,jiang2025detectpointprediction,liu2024cos,gao2024icot}, generating intermediate visual artifacts (e.g., sketches or diagrams) to externalize intermediate states~\citep{hu2024visualsketch, borazjanizadeh2025visualizingthoughtconceptualdiagrams, li2025mvot, ma2025clawmachine}, and hybrid tool-use pipelines that interleave perception and reasoning steps~\citep{visualprogramming,hu2024visualsketch}.

Our work is most closely related to methods that enable \emph{visual re-access} during decoding. We extend the pointer-generator idea~\citep{see2017get} from selective copying in text to selectively reusing visual token embeddings, yielding an interpretable and lightweight mechanism to directly point at the visual evidence.

\section{Visual Grounding Decays During Reasoning}
\label{sec:prelim_exp}

To examine how visual attention evolves at each generation step, we use RefCOCO~\citep{kazemzadeh-etal-2014-referitgame}, a visual reference expression generation benchmark where the task is to create a caption that uniquely identifies a target region specified by a bounding box. Although RefCOCO generally involves shorter generations than mathematical reasoning, it provides a controlled setting for measuring attention dynamics over visual regions. We analyze the TVC-7B model~\citep{sun2025tvc} on the RefCOCO testA split, examining attention from the most recent token to all image tokens across early, middle, and late transformer layers (layers~2,~14, and~27).

\begin{figure}[t]
    \centering
    \includegraphics[width=0.48\linewidth]{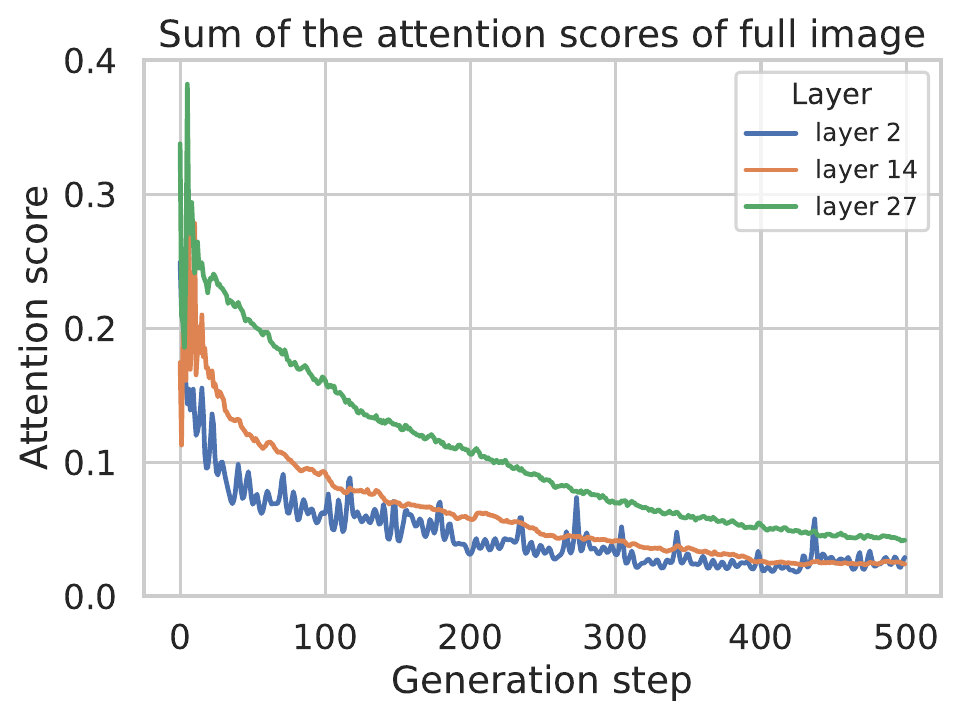}
    \hfill 
    \includegraphics[width=0.48\linewidth]{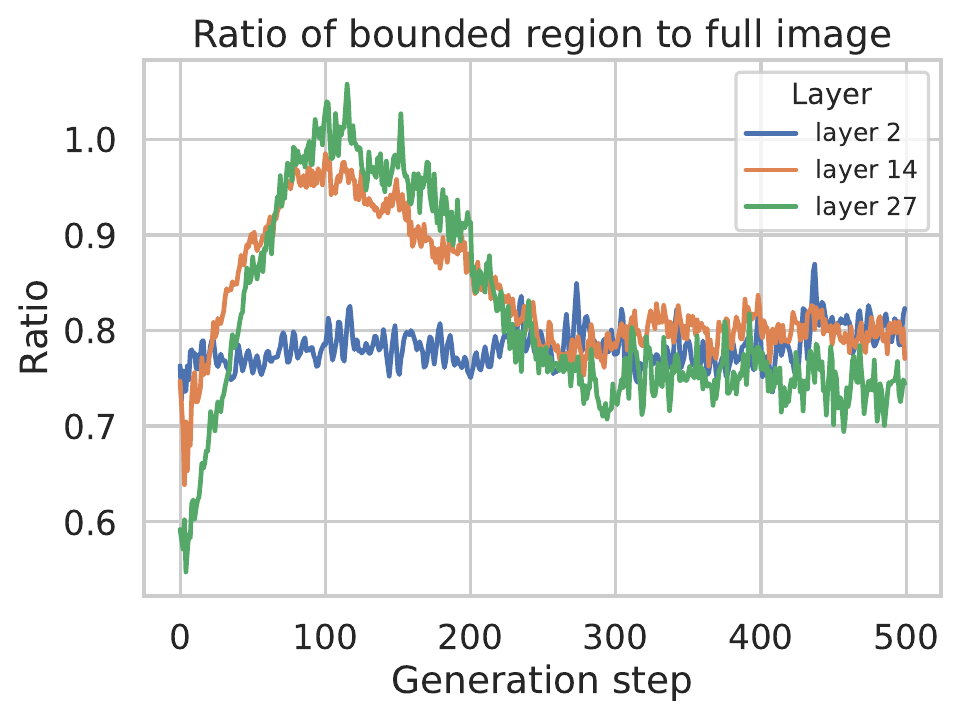}
    
    \caption{\textbf{Left:} Cumulative attention across all visual tokens, showing a gradual decrease in overall attention to the input image tokens. \textbf{Right:} Attention dynamics during reasoning, showing that semantically important visual regions receive disproportionately low attention, suggesting inefficient grounding.}
    \label{fig:attention_analysis}
\end{figure}

\Cref{fig:attention_analysis} left tracks total attention to image tokens and shows a steady decline across decoding steps, indicating a reduced reliance on visual grounding as generation proceeds. 
\cref{fig:attention_analysis} right measures attention to the target region by computing the ratio of mean attention on bounding-box image tokens to that over all image tokens. While layers~14 and~27 initially place greater emphasis on the target region, attention converges to a ratio of $\sim$0.8 by mid-generation, suggesting weakened relative focus on salient visual tokens.

These results suggest \textit{visual grounding decay}, where attention to relevant visual content diminishes during extended generation. This limitation is especially relevant for multimodal reasoning tasks where intermediate steps may benefit from revisiting visual evidence, thereby motivating architectures with dynamic visual access during inference.

\section{Method}
\label{sec:method}

At inference time, \modelname operates as a single, self-contained MLLM: all textual reasoning, visual localization, and point-and-copy operations are performed by the same fine-tuned backbone without invoking external models or tools.

\subsection{Preliminary: Pointing for Language Generation}
\label{subsec:method_prelim}

\paragraph{Formulation.}  
We formulate a conditional next-token prediction objective, as commonly adopted in modern multimodal large language models (MLLMs). Given a sequence of continuous input representations \( c \) (\eg embedded text tokens or visual features) the model is trained to autoregressively predict the discrete next token \( x_t \) conditioned on the input $c$ and previously generated tokens $x_{<t}$:
\[
    p(x_1, \ldots, x_T \mid c) = \prod_{t=1}^T p(x_t \mid c, x_1, \ldots, x_{t-1})
\]

The continuous input sequence \( c \) may include a heterogeneous mixture of modality-specific features, such as embedded discrete text tokens or continuous visual embeddings produced by image encoders (\eg CLIP~\citep{radford2021learningtransferablevisualmodels}). This general formulation covers a wide range of multimodal architectures such as LLaVA~\citep{liu2023llava} and Qwen-VL~\citep{qwen25vl}, which use continuous input image representations.

\paragraph{Pointing.} 
For visually grounded reasoning, rather than generating new visual tokens, we instead teach the model to \textit{point} to its position within the input image sequence it already understands, thereby referencing it explicitly.


The pointing mechanism we examine was first introduced by the pointer-generator network~\citep{see2017get} in text summarization research. 
In the pointer-generator network, the input context sequence $c$ also consists of discrete tokens within the vocabulary space $V$, unlike our setup.
The model dynamically mixes two distributions at each decoding step \( t \): (1) a generation distribution over the vocabulary \( P_{\text{gen}}(x_t) \), and (2) a copy distribution \( P_{\text{ptr}}(x_t) \) over input tokens. The final output probability is given by a gated mixture:
\begin{equation}
\begin{aligned}
p(x_t \mid c, x_{<t}) &=
\lambda_t \, P_{\text{gen}}(x_t)
+ (1-\lambda_t)\, P_{\text{ptr}}(x_t), \\
\lambda_t &\equiv \lambda(x_t \mid c, x_{<t}),
\end{aligned}
\end{equation}

where \( \lambda \in [0,1] \) is a learnable scalar gate that controls the trade-off between generating a new token and copying one from the input.

The pointer distribution is obtained via attention over the encoder representations:
\begin{equation}
\begin{aligned}
\alpha_t^{(k)} &=
\frac{\exp\!\bigl(\mathrm{score}(h_t, c_k)\bigr)}
     {\sum_{k'} \exp\!\bigl(\mathrm{score}(h_t, c_{k'})\bigr)}, \\
P_{\mathrm{ptr}}(x_t = w) &=
\sum_{k:\, w_k = w} \alpha_t^{(k)} .
\end{aligned}
\end{equation}
where \(h_t\) is the decoder hidden state at step \(t\), \(w_k\) the token at position \(k\), and \(\mathrm{score}\) denotes a standard attention scoring function (\eg, dot-product or additive).
We generalize this formulation beyond the original implementation to arbitrary autoregressive language models for explanatory purposes. 

\textbf{Discrete targets.} The above formulation constrains the pointing targets to be within the discrete vocabulary space $V$. This prevents application to general MLLMs as the multimodal inputs often consist of continuous feature sequences~\citep{liu2023llava,qwen25vl}.

\subsection{v1: Pointing for Multimodal Grounded Reasoning}
\label{subsec:method_model}


To overcome these limitations, we introduce \modelname, a lightweight extension to autoregressive MLLMs that enables explicit visual grounding by pointing to continuous input embeddings. \modelname augments the vocabulary with pointer tokens that allow the model to either generate text or copy visual content during inference. All reasoning and grounding are handled within a single fine-tuned backbone (e.g., Qwen2.5-VL), without external modules or auxiliary grounding networks. This design supports unified inference over discrete and continuous modalities without modifying the core architecture (\cref{fig:impl}).

\paragraph{Pointing to continuous inputs.}
The gated mixture formulation of~\cite{see2017get} is unfit for continuous inputs as image embeddings, as such inputs lack discrete mappings to vocabulary tokens \( V \). To enable pointing in this setting, we extend the output space to include references to \emph{positions} in the continuous input. Specifically, we define the augmented output space as \( \bar{V} = V \cup C \), where \( C = \{c_1, c_2, \ldots, c_K\} \) denotes the set of continuous input vectors (e.g. embeddings of the input image patches). This formulation allows the model to generate either a vocabulary token or a pointer to a specific continuous input. We denote a pointer to input vector \( c_k \) as \( \langle \text{ptr}:c_k \rangle \), which is treated as a discrete token during decoding.

At each decoding step \( t \), the model computes two distributions: (1) a generation distribution over the vocabulary \( V \), producing logits \( \text{logit}_{\text{gen}} \in \mathbb{R}^{|V|} \), and (2) a pointing distribution over the input positions \( C \), producing logits \( \text{logit}_{\text{ptr}} \in \mathbb{R}^K \). The final output logits are defined as:
\[
\text{logit}_t = \left[ \text{logit}_{\text{gen}} \,\|\, \text{logit}_{\text{ptr}} \right] \in \mathbb{R}^{|V| + K}
\]
where \( [\cdot \,\|\, \cdot] \) denotes concatenation. Pointing logits are computed by attending over the image sequence:
\[
\text{logit}_{\text{ptr}}^{(k)} = \frac{L_q(h_t) \cdot L_k(c_k)^\top}{\sqrt{D}}
\]
where \( h_t \) is the decoder hidden state at step \( t \), \( L_q \) and \( L_k \) are learned linear projections, and the scaling factor \( \sqrt{D} \) follows standard attention practice. We omit the gating module \( \lambda \) as the logit types are defined over disjoint spaces and do not require interpolation.

During inference, if the model selects an index in \( V \), the next token \( x_t \) is emitted as the corresponding vocabulary token. If the model selects an index \( k \in C \), the token is represented as a pointer \( x_t = \langle \text{ptr}:c_k \rangle \). On the subsequent decoding step, the input embedding at position \( t \) is replaced with the continuous vector \( c_k \), enabling the model to attend directly to the referenced content.

\begin{table*}[t]
  \centering
  \resizebox{1.0\linewidth}{!}{
  \begin{tabular}{l|c|c|cccc|cc}
    \toprule
    \multirow{2}{*}{Model} & \multirow{2}{*}{Size} & Reasoning & MathVista & \multicolumn{2}{c}{MathVision} & MathVerse & \multicolumn{2}{c}{Average} \\
    & & Only & mini & mini & full & mini & mini & full \\
    \midrule
    Qwen2-VL~\citep{qwen2vl}           & 7B   & \xmark & 60.9 & -   & 16.3 & 24.6 & - & 20.5 \\
    Qwen2-VL~\citep{qwen2vl}           & 72B  & \xmark & 69.7 & -   & 26.6 & 36.2 & - & 31.4 \\
    Qwen2.5-VL~\citep{qwen25vl}        & 7B   & \xmark & 67.8 & 23.6 & -   & 44.5 & 45.3 & -    \\
    Qwen2.5-VL~\citep{qwen25vl}        & 72B  & \xmark & 74.8 & 39.8 & -   & 57.6 & 57.4 & -    \\
    InternVL2.5~\citep{internvl25}     & 8B   & \xmark & 64.4 & 22.0 & 19.7 & 39.5 & 41.9 & 29.6 \\
    InternVL2.5~\citep{internvl25}     & 78B  & \xmark & 72.3 & 34.9 & 32.2 & 51.7 & 53.0 & 42.0 \\
    GPT-4o~\citep{gpt4o}               & -    & \xmark & 63.8 & -   & 30.4 & 50.2 & - & 40.3 \\
    \midrule
    LLaVa-CoT~\citep{xu2025llavacotletvisionlanguage} & 11B  & \cmark & 54.8 & 16.3 & -   & 33.9 & 35.0 & -    \\
    Mulberry~\citep{yao2024mulberry}  & 7B   & \cmark & 63.1 & -   & -   & 39.6 & - & -    \\
    TVC~\citep{sun2025tvc}            & 7B   & \cmark & 68.1 & -   & 22.7 & 38.9 & - & 30.8 \\
    TVC~\citep{sun2025tvc}            & 72B  & \cmark & 72.2 & -   & 41.9 & 48.8 & - & 45.4 \\
    QVQ-72B-preview~\citep{qvq-72b-preview} & 72B  & \cmark & 71.4 & 35.9 & -   & 41.5 & 49.6 & -    \\
    \midrule      
    Base (Qwen2.5-VL)        & 7B   & \xmark & 67.8 & 23.6 & -   & 44.5 & 45.3 & -    \\
    Text-Only (TVC)            & 7B   & \cmark & 68.1 & -   & 22.7 & 38.9 & - & 30.8 \\
    \textbf{Ours}                    & 7B   & \cmark & \textbf{68.6} & \textbf{34.5} & \textbf{28.1} & \textbf{48.6} & \textbf{50.6} & \textbf{38.4} \\
    \quad \rotatebox[origin=c]{180}{$\Lsh$} Inference w/o Pointing                & 7B   & \cmark & 60.0  & 25.3 & 23.7 & 33.6 & 39.6 & 28.7 \\
    \bottomrule
  \end{tabular}%
  }
\caption{\textbf{Results on multimodal mathematical reasoning tasks.} We report MathVision on both the mini and full subsets to include additional baselines. Average (mini) is the mean over MathVista (mini), MathVision (mini), and MathVerse (mini); Average (full) is the mean over MathVision (full) and MathVerse (mini), as MathVista has no publicly scorable full split. \textbf{Bold} denotes the best result among 7B-scale models; larger models (72B/78B) and GPT-4o are reported as reference points rather than size-matched competitors. The final block repeats the 7B base and text-only rows alongside \modelname to make the controlled comparison directly readable. Some scores are absent in the references (``-``).}
  \label{tab:main-exp} 
\end{table*}

\subsection{Annotating Visually-grounded Reasoning Data}
\label{subsec:method_data}

To train \modelname, we require fine-grained multimodal reasoning traces in which each step is grounded to specific visual evidence. To this end, we construct \dataname, a dataset of 300K multimodal reasoning paths with interleaved grounding annotations. Each trajectory includes a sequence of reasoning steps, where textual inferences are explicitly linked to corresponding image regions.

We generate the dataset using a fully automated three-stage pipeline: (1) oversampling textual reasoning paths from a pretrained MLLM, (2) decomposing each path into discrete visual queries and retrieval steps via an LLM, and (3) grounding each visual reference by aligning it with a bounding box in the input image (see \Cref{app:data_generation_details,fig:data_pipeline} for details).


\paragraph{Constructing base reasoning traces.}
As a seed to our grounded corpus, we adopt the training set of TVC~\citep{sun2025tvc}, which consists of reasoning traces generated from the QvQ model~\citep{qvq-72b-preview}. 


\paragraph{Decomposing reasoning traces into visual reference steps.}
We use a strong off-the-shelf LLM, Gemini-2.0-flash~\citep{gemini2.0flash}, to extract visual grounding cues from text-based reasoning traces. The model collects and rewrites each visual reference as a \textit{detect} call, which takes a short natural-language description and returns the corresponding image region. Retrieved regions are cached and assigned symbolic identifiers such as \texttt{<objX>} in order of appearance. The LLM also produces a key-value list of visual components, where each key is a unique descriptive reference used in later steps. We guide this process with domain-specific few-shot prompts, described in Appendix~\ref{app:data_generation_details}. We then filter out invalid outputs, including mismatches between references and retrieved objects, non-unique labels, insufficient object count ($\leq 2$), and ill-formed reasoning. After filtering, about 82\% of samples are retained.

\paragraph{Grounding visual references to image regions.}
Visual grounding remains difficult in multimodal reasoning, especially for abstract or unconventional regions such as charts, geometry, and medical scans. Existing grounding models often perform poorly in these settings~\citep{steiner2024paligemma, xiao2024florence}, particularly when the target is abstract or symbolic, such as an angle or geometric relation.

To exploit the implicit visual grounding behavior in MLLMs, we build on Qwen2.5-VL's visual grounding capacity~\citep{qwen25vl}.
However, rather than relying on its coordinate generation interface, we estimate the model's visual focus using a \emph{relative attention mechanism} inspired by~\cite{zhang2025mllms}, in order to better handle non-natural domains and symbolic cues. The corresponding algorithm is detailed in Appendix~\ref{app:visual_grounding_code}. We validate the quality of this pipeline through human evaluation in Appendix~\ref{app:human_eval}.
This produces a curated training set of \textasciitilde300K grounded examples.


\section{Implementation Details}
\label{sec:impl}

\begin{figure*}
  \centering
  \includegraphics[width=1\textwidth]{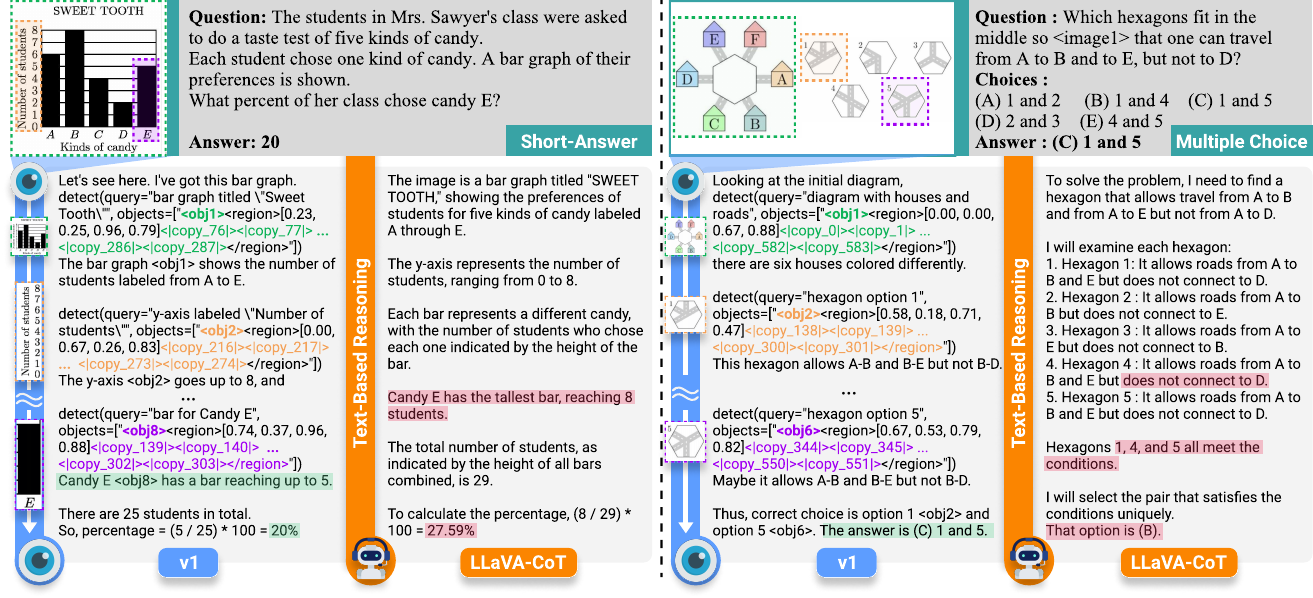}
  \caption{\textbf{Qualitative comparison on MathVision.}  
\modelname's dynamic grounding helps to solve both bar graph and spatial reasoning tasks, while LLaVA-CoT misinterprets visual context.}

  \label{fig:qual1}
\end{figure*}

\paragraph{Preprocessing.}
Given interleaved images, text, and bounding box annotations, we first convert each image into a flattened sequence of patches following the backbone's patchification scheme (e.g., Qwen2.5-VL~\citep{qwen25vl}). Each bounding box is then mapped to a sequence of pointer tokens (e.g., \texttt{<ptr4>}, \ldots, \texttt{<ptr32>}) that index the enclosed patches. These tokens are added to the tokenizer vocabulary without changing the original embedding table or generation head. Instead, during preprocessing, their embeddings are replaced by the corresponding image patch embeddings before entering the transformer. The resulting input is a unified sequence of text tokens, pointer tokens, and image patch embeddings.

\paragraph{Model.}
We instantiate \modelname with Qwen2.5-VL as the MLLM backbone and add only \textit{two lightweight linear layers}: a pointing query head $L_q \in \mathbb{R}^{D \times D}$ and a pointing key head $L_k \in \mathbb{R}^{D \times D}$, where $D$ is the latent dimensionality. Both are initialized as identity matrices scaled by $1/\sqrt{D}$, so their initial effect on the output distribution is minimal. This is well suited to our setting because the pretrained backbone already provides meaningful generative likelihoods $P_{\text{gen}}$, while the pointing module selects at most one position per timestep from $P_{\text{ptr}}$, allowing smooth early training without catastrophic forgetting.

\paragraph{Training.}
Given the dual-nature output space comprising a generative vocabulary \(V\) and a pointing reference set \(K\), we incorporate z-loss regularization to stabilize the softmax partition function, following~\cite{team2024chameleon}. Specifically, we regularize the log-partition function \(Z = \sum_j e^{x_j}\) in the softmax \(\sigma(x)_i = e^{x_i} / Z\) by introducing a z-loss term \(\mathcal{L}_{\text{z}} = \lambda \log \bar{Z}\), where \(\lambda = 10^{-5}\).
To reduce computational overhead, we approximate \(Z\) using a top-\(k=40\) partition function \(\bar{Z} = \sum_{j \in \text{TopK}(x)} e^{x_j}\). This approximation enables efficient and numerically stable training in large-output-space settings. Further details are in~\Cref{sec:ax_resources}.
 

\paragraph{Inference.}
At each decoding step \(x_t\), \modelname uses two extra caches: (1) key features \(L_k(c)\) at image patch positions to compute pointing logits \(\text{logit}_{\text{ptr}}\), and (2) the corresponding image patch features for copying. We implement these by extending the HuggingFace Transformers key-value cache~\citep{wolf-etal-2020-transformers}. The overhead is small, since only a subset of tokens from a single layer is stored.

The added model cost is also minimal: \modelname introduces only one linear projection over image embeddings, negligible relative to the language head. During inference, copied visual tokens increase sequence length, but with revisit suppression they typically stay below 60\% of the text token count, yielding an average total length of about \(1.6\times\) text-only reasoning. No external models or executors are used.

\section{Experiments}
\label{sec:exp}

\subsection{Downstream Evaluation}
\label{subsec:exp_main}

\begin{figure*}
  \centering
  \includegraphics[width=1\textwidth]{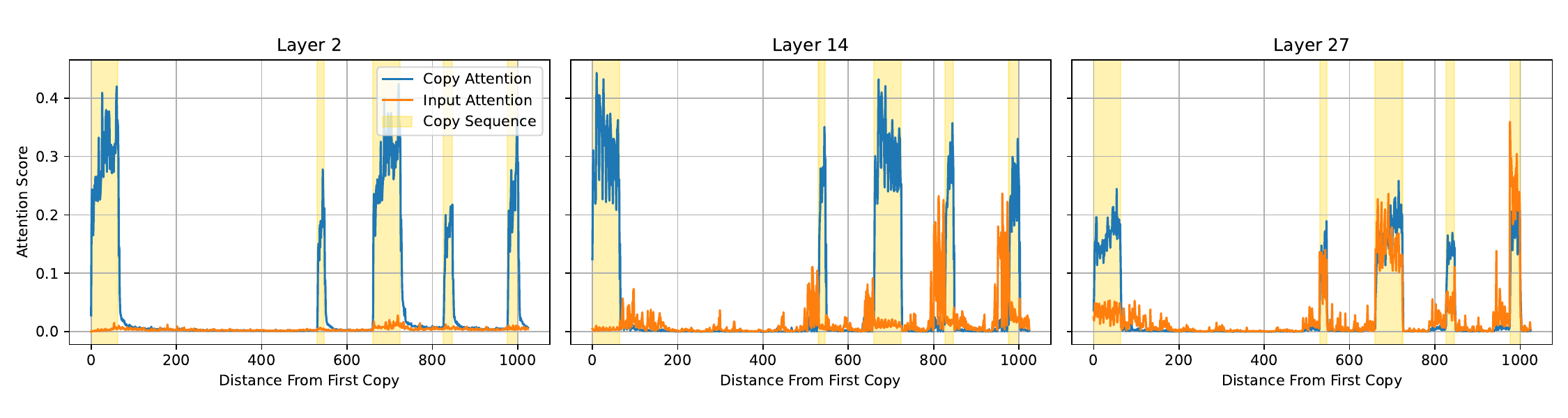}
  \caption{\textbf{Comparison of attention to copy tokens vs. original visual tokens.}  
Layer-wise sum of attention scores directed to copy tokens and their corresponding original visual input tokens from a \modelname. Copy token intervals are shown in \textcolor{yellow!75!black}{yellow}.}

  \label{fig:analysis_attn}
\end{figure*}

\paragraph{Setup.}
We use three representative multimodal mathematical reasoning benchmarks: MathVista (mini)~\citep{lu2024mathvista}, MathVision (mini/full)~\citep{wang2024measuring}, and MathVerse (mini)~\citep{zhang2024mathverse}. Following prior work~\citep{duan2024vlmevalkit}, we compute accuracy with an LLM judge (Gemini-2.0-Flash) that matches predicted and reference answers.

We compare our method against both general-purpose and reasoning-specialized MLLMs. General MLLMs include Qwen2-VL~\citep{qwen2vl} and Qwen2.5-VL~\citep{qwen25vl} at both 7B and 72B scales, as well as InternVL2.5~\citep{internvl25} at 8B and 78B. We also include GPT-4o~\citep{gpt4o} as a high-performing proprietary baseline. For reasoning-oriented models, we evaluate LLaVa-CoT-11B~\citep{xu2025llavacotletvisionlanguage}, Mulberry-7B~\citep{yao2024mulberry}, TVC-7B and 72B~\citep{sun2025tvc}, and QVQ-72B-preview~\citep{qvq-72b-preview}.

\paragraph{Results.}
Quantitative results are presented in~\Cref{tab:main-exp}. Our approach yields substantial performance improvements over baseline models. Among 7B-scale models, \modelname with full pointing capability outperforms both general-purpose and reasoning-specialized baselines. Notably, despite its smaller size, our 7B model narrows the performance gap with several 72B-scale models. The gains are particularly pronounced on MathVision, a benchmark known for its stronger demand for correct grounding.




\subsection{Further Analysis}
\label{subsec:exp_analysis}

\paragraph{Qualitative results.}
\Cref{fig:qual1} compares our method with LLaVA-CoT~\citep{xu2025llavacotletvisionlanguage} on MathVision short-answer and multiple-choice examples. Our \modelname exhibits explicit visual grounding via pointer-based detection and selective copying of relevant image regions.
In the bar graph example, \modelname correctly identifies the bar for Candy~E and computes the correct percentage, while LLaVA-CoT misidentifies the tallest bar. In the hexagon pathfinding task, \modelname reasons over spatial connectivity by attending to structural differences among options, whereas LLaVA-CoT fails to eliminate invalid candidates. These examples illustrate how active visual reference supports more precise and interpretable grounded reasoning than text-only chain-of-thought methods.


\begin{wraptable}{r}{0.48\linewidth}
\vspace{-1.5em}
\centering
\small
\begin{tabular}{l|cc|c|c}
\toprule
Variant & Tr. & Inf. & Score & $\Delta$ \\
\midrule
Backbone           & \xmark & \xmark & 23.6 & --   \\
\midrule
(no pointing) & \cmark & \xmark & 25.3 & +1.7 \\
Ours               & \cmark & \cmark & \textbf{34.5} & \textbf{+10.9} \\
\bottomrule
\end{tabular}
\caption{\textbf{Ablation on MathVision mini.} \textit{Train}: grounded-reasoning training. \textit{Infer}: point-and-copy at inference. $\Delta$: gain relative to the backbone.}
\label{tab:wrap_ablation}
\vspace{-1em}
\end{wraptable}

\paragraph{Ablation study.}
We report an ablation study in~\Cref{tab:wrap_ablation} to isolate the effect of the point-and-copy mechanism. All variants share the same Qwen2.5-VL-7B backbone: \textit{Backbone} applies no task-specific fine-tuning, while \textit{Ours (no pointing)} uses the same training setup as \modelname but disables pointing at inference.
On MathVision (testmini), \modelname improves over the backbone from 23.6 to 34.5 (+10.9), whereas disabling pointing reduces the score to 25.3 (+1.7), indicating that point-and-copy at inference is a key driver of the gains.

For context, TVC~\citep{sun2025tvc} serves as a text-only reasoning baseline; our training data extends its reasoning traces with additional visual-grounding annotations.

\paragraph{How does \modelname utilize pointed visual regions?}
\label{para:attn_score_grounded}
We analyze how \modelname uses visual regions retrieved via the point-and-copy mechanism. As shown in \Cref{fig:analysis_attn}, we compare \textit{Input Attention} (to original visual tokens) and \textit{Copy Attention} (to copied tokens) during generation after the first copy operation.

We observe a structured sequence of behaviors. Before copying, attention to original image tokens increases, indicating a localization step in which relevant regions are identified. Immediately after copying, intermediate layers (e.g., layers~2 and~14) exhibit dominant copy attention, reflecting focused post-retrieval processing of the retrieved region. When averaged across layers, copied tokens consistently receive higher attention than original image tokens, suggesting that copied regions serve as stable and accessible references. In higher layers (e.g., layer~27), attention to input and copied tokens becomes more balanced, which may correspond to a late-stage integration of retrieved visual information into the broader reasoning context.

\section{Conclusion}
\label{sec:con}


We introduced \modelname, a lightweight extension that enables MLLMs to actively revisit input images via a point-and-copy mechanism, and \dataname, a dataset of 300K multimodal reasoning traces with fine-grained visual grounding. Experiments on established multimodal mathematical reasoning benchmarks show that \modelname improves performance, particularly on tasks requiring grounded, multi-step visual reasoning. We hope this work encourages further exploration of dynamic visual access as a core component of multimodal reasoning.

\ifcolmfinal
\section*{Acknowledgements}

This work was partly supported by the Institute of Information \& Communications
Technology Planning \& Evaluation (IITP) grant funded by the Korean government
(MSIT) (No.~RS-2021-II211343, Artificial Intelligence Graduate School Program
(Seoul National University), No.~RS-2026-25522885, Development of a World
Foundation Model for Training and Development of Physical AI Systems,
No.~RS-2026-25512061, Development of Large Action Model-Based Autonomous Digital
Twin Operation Technology for Proactive Problem Solving), the National Research
Foundation of Korea (NRF) grant funded by the Korean government (MSIT)
(No.~RS-2024-00354218, No.~RS-2026-25561904), and the Technology Innovation
Program (RS-2025-25456760, Development of a humanoid robot specialized in chemical
processes based on AI foundation model) funded by the Ministry of Trade, Industry
and Resources (MOTIR, Korea). We express special thanks to KAIT GPU project. The
ICT at Seoul National University provides research facilities for this study.

\fi

\bibliography{references}
\bibliographystyle{colm2026_conference}

\newpage
\appendix
\onecolumn
\appendix

\section*{Overview of the Appendix}
\label{sec:ax_overview}

This Appendix is structured as follows:
\begin{itemize}[leftmargin=*]
    \item \Cref{sec:ax_resources} describes implementation details and resources used in the project;
    \item \Cref{sec:ax_additional} reports additional experiments on generalization, cross-backbone transfer, test-time scaling, and training compute;
    \item \Cref{sec:lim} discusses limitations and future directions;
    \item \Cref{app:data_generation_details} provides details of the data generation process;
    \item \Cref{app:human_eval} reports human evaluation results validating the grounding quality of our training dataset (\dataname) and our model (\modelname);
    \item \Cref{app:visual_grounding_code} details pseudo-code on the visual grounding pipeline we utilized in the data generation process;
    \item \Cref{app:use_of_llms} presents the use of Large Language Models (LLMs);
    \item \Cref{app:additional_qual_results} presents additional qualitative results.
\end{itemize}

\section{Implementation Details \& Resources}
\label{sec:ax_resources}

\paragraph{Training Details}
All models are trained under uniform settings: a base learning rate of \( 3 \times 10^{-5} \), per-device batch size of 2, and gradient accumulation over 4 steps. We leverage DeepSpeed for distributed training across 8 NVIDIA A100 GPUs. Optimization uses AdamW with \( \beta_1 = 0.9 \), \( \beta_2 = 0.95 \), and training is performed for 5 epochs. Training \modelname requires approximately 672 A100-equivalent GPU-hours (300K examples over 5 epochs on 8 NVIDIA A100 GPUs).

\paragraph{Inference Cost}
\Cref{tab:latency} reports measured inference cost per sample under HuggingFace \texttt{.generate} and batched vLLM inference. Point-and-copy adds roughly 990 copied visual tokens per sample, yet the wall-clock overhead relative to the text-only TVC-7B baseline is modest, about 7\% under both runners. The \(1.6\times\) sequence-length figure quoted in the main text is a conservative upper bound derived from training-data statistics rather than a measured runtime cost.

\begin{table}[t]
  \centering
  \caption{\textbf{Inference cost per sample.} Total tokens, copied visual tokens, wall-clock time, and throughput under HuggingFace \texttt{.generate} and batched vLLM. Dashes indicate quantities not separately instrumented under batched inference.}
  \label{tab:latency}
  \small
  \begin{tabular}{llrrrr}
    \toprule
    Runner & Model & Tokens & Copy tokens & Wall clock & Tokens/sec \\
    \midrule
    HF \texttt{.generate} & TVC-7B & 4{,}057 & 0 & 143.42\,s & 28.29 \\
    vLLM (batched) & TVC-7B & -- & -- & 23.86\,s & 110.38 \\
    HF \texttt{.generate} & \modelname-7B & 4{,}087 & 990 & 154.04\,s & 26.53 \\
    vLLM (batched) & \modelname-7B & -- & -- & 25.65\,s & 103.24 \\
    \bottomrule
  \end{tabular}
\end{table}

\paragraph{Training Duration}
Our training schedule of 5 epochs follows the setup used for the original text-only reasoning trace dataset, which we extend to the grounded reasoning setup~\citep{sun2025tvc}. Because the reasoning traces contain substantially longer token sequences than typical MLLM data, shorter training runs produced unstable behaviors such as repetition and incomplete reasoning without a final answer. In contrast, the point-and-copy behavior required relatively little data and typically saturated within the first epoch, as indicated by an early plateau in copy-token accuracy. The longer schedule therefore reflects the requirements of the inherited reasoning-trace setup rather than the needs of the pointer mechanism itself.


\section{Additional Experimental Results}
\label{sec:ax_additional}

\begin{figure}[t]
  \centering
  \includegraphics[width=\linewidth]{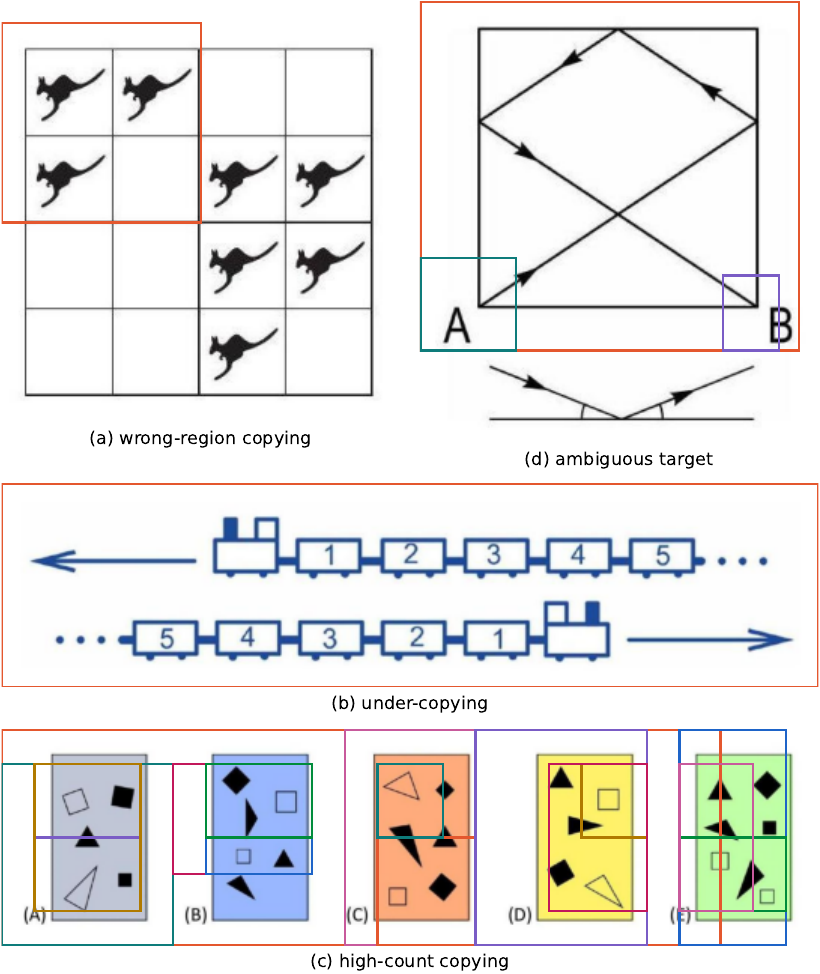}
  \caption{\textbf{Representative failure modes of \modelname{}} on MathVision. Boxes show the regions selected by the model, colored by step. (a) The query names the full $8\times8$ grid but only its top-left quadrant is selected. (b) A single whole-image selection, after which reasoning proceeds without further visual access. (c) Sixteen pointer calls across answer options and their sub-shapes. (d) A reflected billiard path, which no axis-aligned box can isolate.}
  \label{fig:failure}
\end{figure}

This section reports supplementary experiments that probe how far the point-and-copy mechanism generalizes: beyond mathematics (\cref{tab:emma}), beyond a single backbone (\cref{tab:crossbackbone}), against sequential test-time scaling (\cref{tab:tts}), and relative to the training compute of reinforcement-learning alternatives (\cref{tab:compute}). We also analyze where the mechanism still fails (\cref{fig:failure}).

\paragraph{Generalization beyond mathematics.}
Although \dataname{} is built for mathematical grounded reasoning, the point-and-copy mechanism itself is domain-agnostic. We therefore evaluate on two non-math subsets, physics and coding, of the EMMA-mini multimodal reasoning benchmark~\citep{hao2025emma}. As shown in \cref{tab:emma}, \modelname{} improves over both the Qwen2.5-VL-7B backbone and the matched no-copy variant on both domains, suggesting the gain is connected to explicit visual re-access rather than to math-specific supervision alone.

\begin{table}[t]
  \centering
  \caption{\textbf{Generalization beyond mathematics (EMMA-mini).} Accuracy (\%) on the physics and coding subsets. Rows below the rule are 7B-scale models evaluated in our pipeline; rows above are larger reference models.}
  \label{tab:emma}
  \begin{tabular}{lcc}
    \toprule
    Model & Physics & Coding \\
    \midrule
    GPT-4o & 44.0 & 38.0 \\
    Qwen2-VL-72B & 40.0 & 37.0 \\
    \midrule
    Qwen2.5-VL-7B & 30.0 & 32.0 \\
    \modelname{} w/o copying & 31.0 & 28.0 \\
    \modelname{} & \textbf{36.0} & \textbf{40.0} \\
    \bottomrule
  \end{tabular}
\end{table}

\paragraph{Cross-backbone transfer.}
To test whether the gains depend on a particular backbone, we apply point-and-copy to Gemma3-4B~\citep{gemmateam2025gemma3} while keeping the grounding supervision unchanged. \Cref{tab:crossbackbone} shows consistent improvements across all three math benchmarks, indicating that the benefit is driven by the mechanism rather than by backbone-specific attention behavior.

\begin{table}[t]
  \centering
  \caption{\textbf{Cross-backbone transfer.} Applying point-and-copy to Gemma3-4B (\modelname-Gemma3) improves over the vanilla backbone across math benchmarks. Gemma3-4B numbers are the OpenCompass leaderboard reference.}
  \label{tab:crossbackbone}
  \begin{tabular}{lccc}
    \toprule
    Benchmark & Gemma3-4B & \modelname-Gemma3 & $\Delta$ \\
    \midrule
    MathVision-mini & 21.05 & 25.00 & +3.95 \\
    MathVista-mini & 46.30 & 60.40 & +14.10 \\
    MathVerse-mini & 29.54 & 37.80 & +8.26 \\
    \bottomrule
  \end{tabular}
\end{table}

\paragraph{Comparison to sequential test-time scaling.}
A natural question is whether simply spending more decoding budget on a text-only model recovers the benefit of visual re-access. Because a trained \modelname{} cannot be scaled down at test time without truncating its answer, we apply sequential test-time scaling~\citep{muennighoff2025s1} to the text-only TVC-SFT baseline instead. \Cref{tab:tts} shows that increasing the text-only budget from 8{,}192 to 12{,}800 tokens yields only a marginal gain and remains far below \modelname{} at the same budget.

\begin{table}[t]
  \centering
  \caption{\textbf{Sequential test-time scaling on MathVision-mini.} Text-only chain-of-thought at matched token budgets versus \modelname{}.}
  \label{tab:tts}
  \begin{tabular}{llcc}
    \toprule
    Model & Size & Token budget & Score \\
    \midrule
    Text-only & 7B & 8{,}192 & 24.0 \\
    Text-only & 7B & 12{,}800 & 24.7 \\
    \modelname{} & 7B & 12{,}800 & \textbf{34.5} \\
    \bottomrule
  \end{tabular}
\end{table}

\paragraph{Failure cases.}
Point-and-copy provides explicit visual re-access, but it does not eliminate reasoning errors. Inspecting incorrect MathVision predictions, we observe four recurring failure modes, shown in \cref{fig:failure}. In \emph{wrong-region copying} (a), the pointed region does not contain the queried object: the model asks for an $8\times8$ grid but points only at its top-left quadrant. In \emph{under-copying} (b), the model points once at the entire image and then continues in text without re-accessing localized evidence. In \emph{high-count copying} (c), the question requires tracking many distinct instances, and the model issues sixteen separate pointer calls across the answer options without resolving the count. In \emph{ambiguous targets} (d), the referent cannot be delimited by a box at all, such as the reflected path of a ball on a billiard table. These cases indicate that point-and-copy improves access to visual evidence but does not by itself resolve localization ambiguity or long-horizon symbolic reasoning.

\paragraph{Training compute relative to RL-based alternatives.}
\modelname{} is trained with supervised fine-tuning rather than online policy optimization, so reinforcement-learning methods such as GRPO are not a like-for-like baseline: they change both the optimization objective and the source of supervision. They are nonetheless a useful compute reference. Wall-clock cost is rarely reported for VL-GRPO runs, so in \cref{tab:compute} we compile runs with reported or estimable compute and normalize them to A100-equivalent GPU-hours using published BF16 H100/A100 throughput ratios of roughly $2.0$--$2.3\times$.

\begin{table}[t]
  \centering
  \caption{\textbf{Training compute reference.} A100-equivalent GPU-hours for \modelname{} and for reinforcement-learning vision-language runs with reported or estimable compute. VL-Rethinker additionally uses a Forced-Rethinking SFT stage. GRPO entries are a compute reference, not a controlled baseline.}
  \label{tab:compute}
  \small
  \begin{tabular}{lllr}
    \toprule
    Method & Setting & Training data & A100-equiv.\ GPU-hrs \\
    \midrule
    \modelname{} (ours) & 7B-VL SFT & 300K $\times$ 5 epochs & 672 \\
    Visionary-R1~\citep{xia2025visionaryr1} & 3B-VL GRPO & 272.6K & ${\sim}1{,}500$ \\
    VL-Rethinker~\citep{wang2025vlrethinker} & 7B-VL GRPO & 16K $\times$ 3 epochs & ${\sim}2{,}000$--$2{,}300$ \\
    VLM-R1~\citep{shen2025vlmr1} & 3B-VL GRPO & 2 epochs & $3{,}200$--$3{,}700+$ \\
    VLM-R1~\citep{shen2025vlmr1} & 3B-VL GRPO & extended run & $11{,}200$--$12{,}900+$ \\
    \bottomrule
  \end{tabular}
\end{table}

\section{Limitations and Future Work}
\label{sec:lim}

This work focuses on demonstrating the effectiveness of active visual reference in structured multimodal reasoning via a simple point-and-copy mechanism. While \modelname shows strong performance in mathematical domains, several directions remain for broader applicability.

\paragraph{Beyond mathematical domains.} Extending \modelname to other settings (\eg, scientific diagrams, medical images, or visual commonsense) presents new challenges in representation and supervision. These domains often lack structured reasoning traces, making data collection more difficult. Since \dataname relies on a pretrained text-only MLLM to seed reasoning, generalizing to less structured domains will require advances in decomposition, grounding, and alignment.

\paragraph{Weak supervision and reinforcement learning.} Recent work in inference-time scaling and alignment has shown the promise of reward-based learning for reasoning. Incorporating such methods into \modelname may enable more flexible and efficient visual retrieval strategies without dense supervision. We leave this exploration to future work due to current resource constraints.

\section{Data Generation Details}
\label{app:data_generation_details}

\begin{figure}[!htbp]
 \centering
 \includegraphics[width=1\columnwidth]{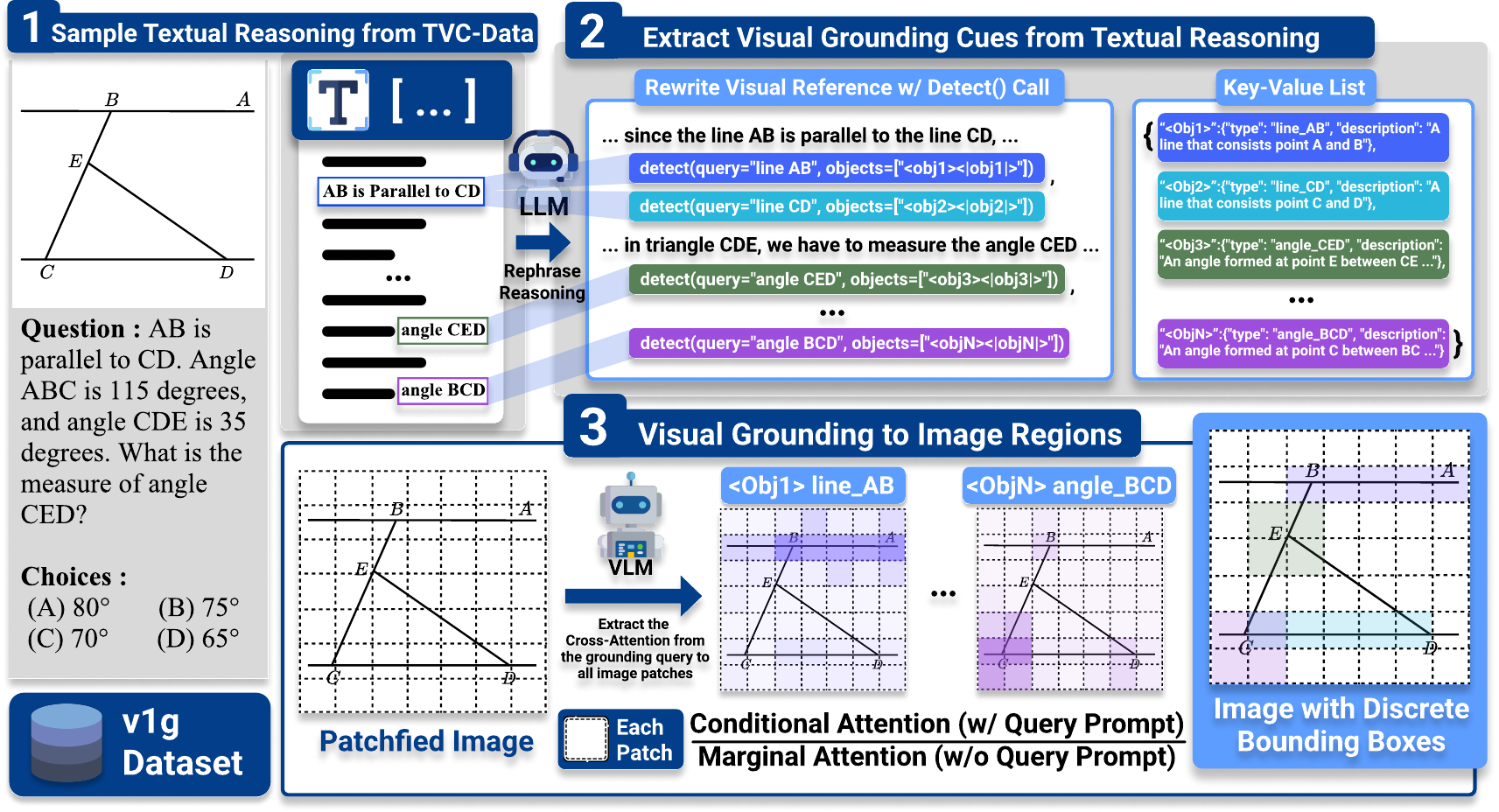}
 \caption{\dataname dataset construction pipeline.}
 \label{fig:data_pipeline}
\end{figure}

\Cref{fig:data_pipeline} illustrates the construction pipeline for our \dataname dataset; each stage of this pipeline is described in detail in \Cref{subsec:method_data}. 
The specific prompt template used to decompose text-based reasoning paths into visual queries (as outlined in our methodology in \Cref{subsec:method_data}) is provided in \Cref{tab:app_data_generation_prompt}. 

\section{Human Evaluation of Grounding Quality}
\label{app:human_eval}

\subsection{Evaluation of v1g Dataset Quality}

To validate the quality of our automatically generated visual grounding annotations in the v1g dataset, we conducted a human evaluation comparing our attention-based grounding approach against GroundingDINO~\citep{liu2024grounding}, a widely used open-set object detector.

\paragraph{Methodology.} 
We randomly sampled 100 examples from the v1g dataset, each containing multiple bounding boxes. Three expert annotators independently evaluated each bounding box using binary classification on three criteria:
\begin{itemize}
    \item \textbf{Correctness}: Whether the bounding box covers the intended object or region
    \item \textbf{Comprehensiveness}: Whether all relevant visual content is included within the box
    \item \textbf{Tightness}: Whether the box is well-fitted with minimal extraneous background
\end{itemize}

We report the average score across annotators, majority vote, and Fleiss' $\kappa$ to assess inter-annotator agreement. Agreement quality follows standard interpretations: Fair (0.21--0.40), Moderate (0.41--0.60), Substantial (0.61--0.80), and Almost Perfect (0.81--1.00).

\paragraph{Results.} As shown in Table~\ref{tab:dataset_eval}, our attention-based grounding method substantially outperforms GroundingDINO on correctness (83.3\% vs. 29.3\% average score), demonstrating superior capability in localizing semantically complex and context-dependent entities such as geometric elements (e.g., ``angle ABC''), chart components (e.g., ``bar for Grace''), and referring expressions (e.g., ``the second figure''). While GroundingDINO achieves higher inter-annotator agreement, this primarily reflects consistent failure modes rather than quality, as evidenced by its low absolute performance.

\begin{table}[t]
\centering
\caption{Human evaluation of v1g dataset quality.}
\label{tab:dataset_eval}
\small
\begin{tabular}{llcccc}
\toprule
\textbf{Method} & \textbf{Metric} & \textbf{Avg} & \textbf{Majority} & \textbf{Fleiss' $\kappa$} & \textbf{Agreement} \\
\midrule
\multirow{3}{*}{Attention-based (Ours)} 
& Correctness & 83.3\% & 87.0\% & 0.352 & Fair \\
& Comprehensive. & 55.0\% & 56.0\% & 0.582 & Moderate \\
& Tightness & 46.0\% & 44.0\% & 0.436 & Moderate \\
\midrule
\multirow{3}{*}{\shortstack[l]{Grounding-\\DINO}} 
& Correctness & 29.3\% & 30.0\% & 0.711 & Substantial \\
& Comprehensive. & 47.3\% & 49.0\% & 0.906 & Almost Perfect \\
& Tightness & 19.0\% & 18.0\% & 0.675 & Substantial \\
\bottomrule
\end{tabular}
\end{table}

\subsection{Evaluation of v1 Pointing Accuracy}

We additionally evaluated the pointing accuracy of our trained v1 model to assess how effectively it grounds visual references during inference.

\paragraph{Methodology.} Using the same evaluation protocol, we sampled 100 outputs from v1 on the MathVision dataset. Annotators evaluated whether the model's pointed regions (copied image tokens) correctly corresponded to the referenced objects in the reasoning trace. We added an \textbf{Appropriateness} criterion to assess whether the pointing action was contextually justified.

\paragraph{Results.} Table~\ref{tab:model_eval} demonstrates that v1 maintains high grounding quality during inference, achieving 82.7\% correctness; comparable to the training data quality. The high appropriateness score (87.7\%) indicates that the model learns to selectively invoke the pointing mechanism when dynamic visual reference is genuinely beneficial for reasoning.

\begin{table}[t]
\centering
\caption{Human evaluation of v1 model pointing quality.}
\label{tab:model_eval}
\small
\begin{tabular}{lcccc}
\toprule
\textbf{Metric} & \textbf{Avg} & \textbf{Majority} & \textbf{Fleiss' $\kappa$} & \textbf{Agreement} \\
\midrule
Correctness & 82.7\% & 87.0\% & 0.558 & Moderate \\
Comprehensiveness & 55.7\% & 54.0\% & 0.689 & Substantial \\
Tightness & 49.3\% & 40.0\% & 0.280 & Fair \\
Appropriateness & 87.7\% & 90.0\% & 0.599 & Moderate \\
\bottomrule
\end{tabular}
\end{table}



\paragraph{Discussion.} The evaluation reveals that our attention-based grounding excels at capturing semantically rich visual references that are challenging for traditional object detectors. The moderate tightness scores across both methods reflect the inherent ambiguity in defining precise boundaries for abstract concepts (e.g., "angle 2" or "the second figure"), where multiple valid interpretations exist. The consistency between training data quality and model performance suggests that v1 successfully learns robust visual grounding capabilities from our automatically generated supervision.

\section{Bounding-Box Extraction from Cross-Attention}
\label{app:visual_grounding_code}

This section provides a high-level pseudocode description of our data annotation method for deriving
bounding boxes from cross-attention in Qwen2.5-VL.

\begin{algorithm}
\caption{Bounding-Box Extraction from Cross-Attention (High-Level)}
\begin{algorithmic}
\STATE {\bfseries Input:} Image $I$, region description $T$
\STATE {\bfseries Output:} Bounding box $b$ corresponding to $T$
\STATE
\STATE \textbf{1. Prepare multimodal input.}
\STATE Concatenate $I$ with a static visual-grounding instruction prompt and feed it to Qwen2.5-VL.
\STATE
\STATE \textbf{2. Extract attention with instruction.}
\STATE From the final decoding position, obtain the cross-attention map $A$ over image tokens. Use a predefined set of layers (selected empirically) and average across heads.
\STATE
\STATE \textbf{3. Extract baseline attention.}
\STATE Remove the object name from the prompt, feed the modified prompt with $I$ to the model, and extract the corresponding attention map $A'$ using the same layers and averaging.
\STATE
\STATE \textbf{4. Compute attention contrast.}
\STATE Compute the contrastive relevance for each image token: $R = A / A'$.
\STATE
\STATE \textbf{5. Derive bounding region.}
\STATE Identify the peak region in $R$. Sweep over multiple candidate crop ratios; for each ratio, form a bounding region around the peak. Select the bounding box maximizing contrast sharpness between inside and outside regions. Convert the selected region to image-coordinate bounding box $b$.
\STATE
\STATE \textbf{6. Return.}
\STATE \textbf{return} $b$
\end{algorithmic}
\end{algorithm}

\section{Use of LLMs}
\label{app:use_of_llms}
This work uses LLMs for data generation (Gemini-2.0-Flash for reasoning trace decomposition) and evaluation (Gemini-2.0-Flash for answer matching), as described in Sections 4.3 and 6.1.

\section{Additional Qualitative Results}
\label{app:additional_qual_results}

To further illustrate \modelname's complex visual reasoning, this section provides additional qualitative examples, complementing \Cref{fig:qual1} from the main text. These examples highlight how \modelname leverages the point-and-copy mechanism.

\begin{figure}
  \centering
  \includegraphics[width=0.8\columnwidth]{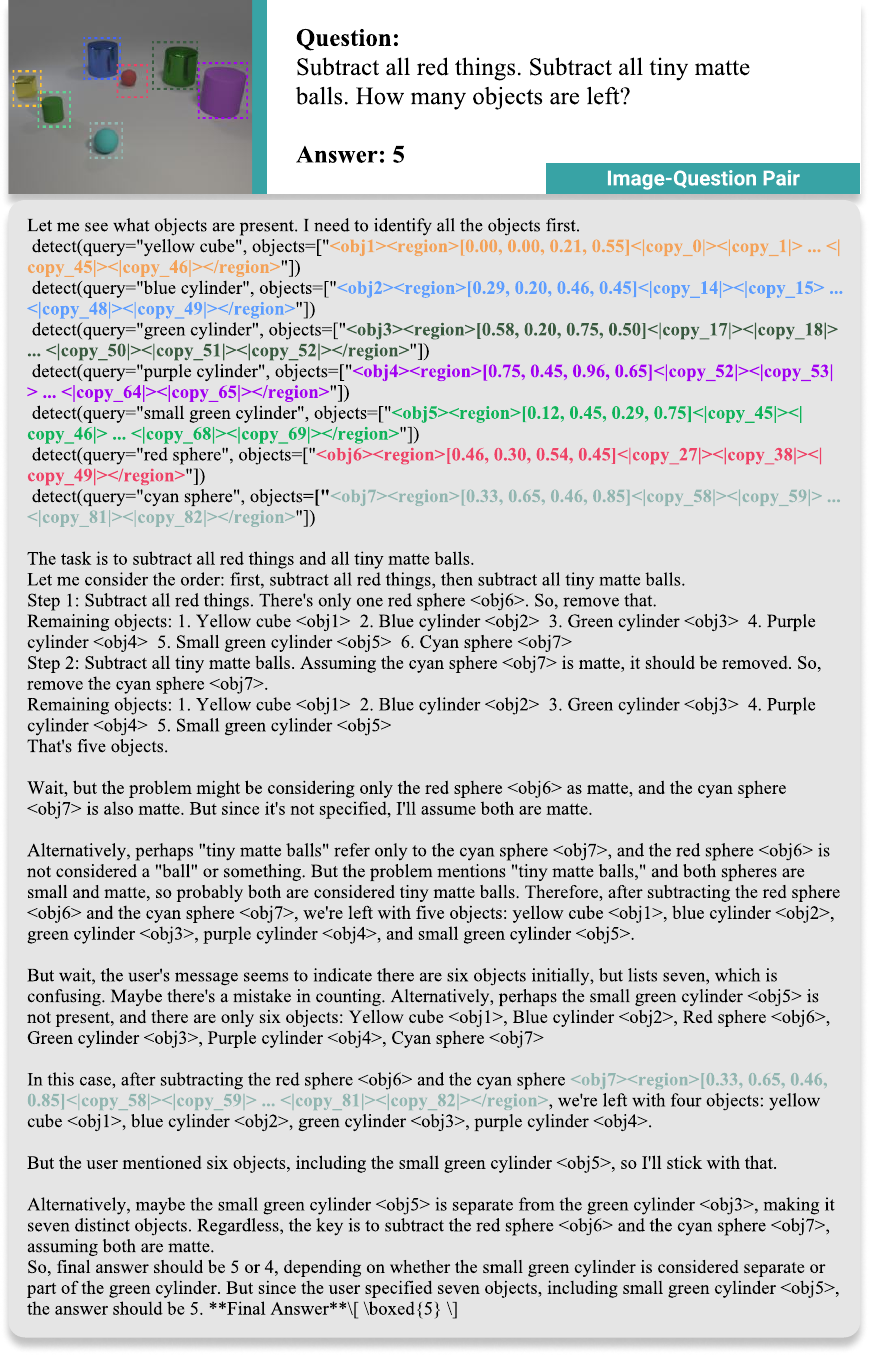}
  \caption{Qualitative example of \modelname tackling an attribute-based counting task in a synthetic domain.}
  \label{fig:app_qual_example1}
\end{figure}

\Cref{fig:app_qual_example1} demonstrates \modelname on a synthetic task (CLEVR-like) requiring object counting based on the query: ``Subtract all red things, then subtract all tiny matte balls. How many objects are left?''. \modelname first localizes objects using its pointer mechanism. It then sequentially reasons, identifying ``red'' objects before revisiting relevant items, like the ``cyan sphere,'' to verify the combined ``tiny'' and ``matte'' attributes through targeted attention. This process demonstrates \modelname's capacity for precise attribute grounding and multi-step compositional reasoning enabled by the point-and-copy mechanism.

\begin{figure}
  \centering
  \includegraphics[width=0.8\columnwidth]{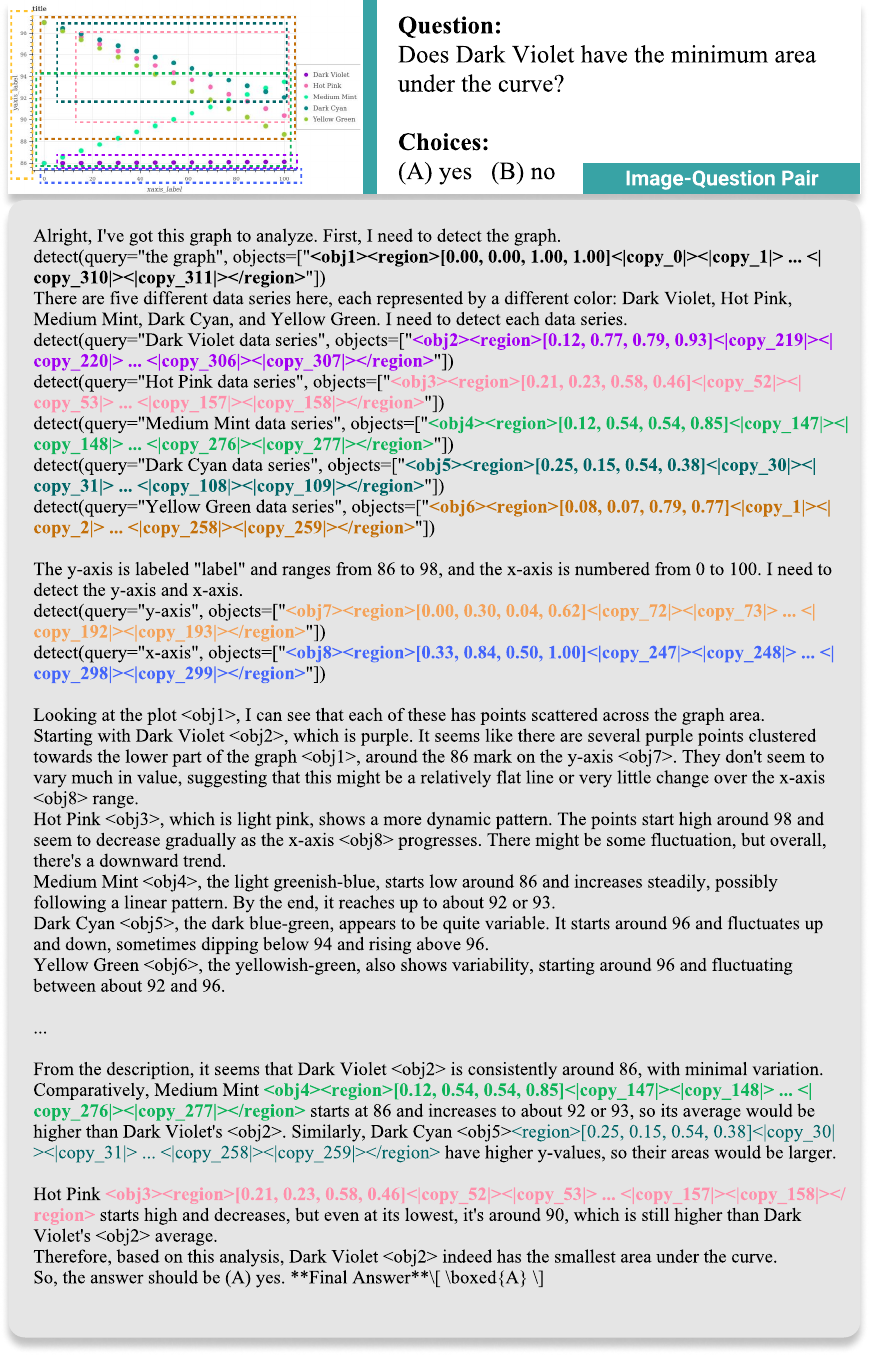}
  \caption{Qualitative example of \modelname performing comparative reasoning on a chart comprehension task.}
  \label{fig:app_qual_example2}
\end{figure}

In~\Cref{fig:app_qual_example2}, \modelname tackles a chart comprehension task: determining if the ``Dark Violet'' data series has the minimum area under the curve. \modelname initially grounds key chart elements, using its pointer to isolate data series such as ``Dark Violet,'' ``Medium Mint,'' and ``Dark Cyan.'' Later in its reasoning, it proactively revisits these series, performing a comparative analysis of their visual trajectories and relative y-axis values to infer their respective areas. Such selective re-focusing showcases its ability to perform nuanced comparisons within dense visual information.

These examples further affirm that \modelname's architecture, by supporting active visual reference and precise grounding via its pointing mechanism, achieves robust, interpretable, and accurate multi-step visual reasoning.

\clearpage

\clearpage

\clearpage
\begin{tcolorbox}[breakable, title=Prompt for data generation]
You are given text-only reasoning for visual question answering. \\
Your task is to convert this text-only reasoning into visually grounded reasoning.\\
\#\#\# STEP-BY-STEP INSTRUCTION\\
Please follow these instructions step-by-step, imitating human visual reasoning behavior by:\\
1) Start from the beginning of the reasoning and read EACH sentence.\\
2) When you think you'd better look at the object or region, use the detect() function.\\
3) Format: `detect(query="visual item that you want to find", objects=["<obj\#>"])`  \\
4) After detection, reference the visual element with '<obj\#>' tags every time you need to look at it again immediately after mentioning the item.  \\
5) Use NEW object numbers (`<obj1>`, `<obj2>`, `<obj3>`...) for EACH new detection.  \\
\\
\#\#\# EXAMPLE:  \\
Original text:  \\
"Looking at the graph, I can see the function reaches its maximum at x = 3."\\
\\
Corrected:  \\
```\\
To answer the question, I need to look the graph.\\
detect(query="function graph", objects=["<obj1>"])\\
Looking at the graph <obj1>, I can see the function reaches its maximum at x = 3.\\
```\\
Later reference:\\
You can skip the <obj\#> tag when you think you do not need to look it again.\\
```\\
The slope of the function becomes zero at this point on the graph.\\
```\\
\\
\#\#\# KEY REQUIREMENTS:\\
- Every item in lists MUST have its own `detect()` statement  \\
- Put `detect()` statements on their own lines  \\
- NEVER skip any visual element mentioned in the reasoning  \\
- Start object numbering at `obj1` and increase by 1 for each new object  \\
\\
\#\#\# <OBJ\#> REQUIREMENTS\\
- Visual element should be concrete, distinct, and explicit. Later you will localize the element based on the detect(). So make sure that the element not confusing.\\
- Use separate tags for each object (write "between the bus <obj1> and the car <obj2>" not "between <obj1 and obj2>").\\
- GOOD grounding: "I need to analyze this problem. detect(query="triangle", objects=["<obj1>"]) The triangle <obj1> has a right angle at vertex S."\\
- BAD grounding: "detect(query="triangle and rectangular", objects=["<obj1>"]) in the diagram, there are the triangle and rectangular has a right angle." (referring to non-atomic element)\\
- BAD grounding: "detect(query="region", objects=["<obj1>"]) The triangle <obj1> has a right angle." (referring to ambiguous element)\\
\\
After completing the reasoning, list all objects detected:\\
\{\\
  "obj1": \{"type": "function\_graph",\\ "description": "Graph of a function with maximum at x = 3"\},\\
  "obj2": \{"type": "next\_item", \\"description": "Description of next item"\}\\
\}\\
\\
- We will localize the element with the open-world detector based on the description, so make sure to include well-described full self-contained description enough to uniquely identify the object.\\
\\
\#\#\# FINAL FORMAT:\\
\{\\
    "reasoning": "Your fully visually-grounded reasoning text",\\
    "obj\_list": "Your JSON object list"\\
\}\\
\\
Now, strictly following the instruction and the example, please provide the object list and visually grounded reasoning for the following prompt and reasoning:\\
\\
\#\#\# Example\\
Original Conversation\\
\\
HUMAN:\\
{[}few\_shot\_question{]}\\
\\
GPT:\\
{[}few\_shot\_answer{]}\\
\\
\\
\#\#\# Visually Grounded Reasoning\\
\\
GPT: \\
{[}few\_shot\_reasoning{]}\\
\\
\#\#\# Object List:\\
\\
{[}few\_shot\_objects{]}\\
\\
\\
\\
Now, given the conversation, please convert GPT's text-only reasoning into visually grounded reasoning\\
\\
Original Conversation:\\
\\
HUMAN:\\
{[}question{]}\\
\\
GPT:\\
{[}answer{]}\\
\\
\#\#\# Visually Grounded Reasoning:\\
\end{tcolorbox}
\captionof{table}{Prompt used for converting textual reasoning to grounded reasoning annotations in \dataname data generation process.}
\label{tab:app_data_generation_prompt}

\end{document}